# Highly Accurate Robot Calibration Using Adaptive and Momental Bound with Decoupled Weight Decay

Tinghui Chen Student Member, *IEEE* and Shuai Li, *Senior Member, IEEE*

*Abstract*—Within the context of intelligent manufacturing, industrial robots have a pivotal function. Nonetheless, extended operational periods cause a decline in their absolute positioning accuracy, preventing them from meeting high precision. To address this issue, this paper presents a novel robot algorithm that combines an adaptive and momental bound algorithm with decoupled weight decay (AdaModW), which has three-fold ideas: a) adopting an adaptive moment estimation (Adam) algorithm to achieve a high convergence rate, b) introducing a hyperparameter into the Adam algorithm to define the length of memory, effectively addressing the issue of the abnormal learning rate, and c) interpolating a weight decay coefficient to improve its generalization. Numerous experiments on an HRS-JR680 industrial robot show that the presented algorithm significantly outperforms state-of-the-art algorithms in robot calibration performance. Thus, in light of its reliability, this algorithm provides an efficient way to address robot calibration concerns.

## I. INTRODUCTION

Industrial robots have advantages like fast operation speed, high efficiency, and flexible control systems making them increasingly significant in intelligent manufacturing [1]-[16]. However, the demand for operational accuracy and work efficiency of robots is increasing as the applications of industrial robots continue to expand [19]-[38]. The robot positioning accuracy is affected by various factors, including the machining assembly, elastic or inelastic deformation, wear, collisions, etc. The total errors of a robot are composed of kinematic errors and dynamic errors, and the dynamic errors constitute less than 10% [39]-[[51]. Thus, we concentrate on addressing the kinematic errors through kinematic calibration.

Aiming to achieve high-efficiency kinematic error calibration for robots, pioneering scholars have conducted numerous in-depth research [52]-[63]. For instance, Kwon et al. [64] identify the robotic kinematic perimeters with the least square (LS) algorithm, which achieves a significant reduction in model bias. Fin *et al.* [65] build a robot kinematic error model based on an extended Kalman filter (EKF) algorithm. Empirical studies implemented on a novel 5-DOF hybrid robot demonstrate that the EKF algorithm evidently outperforms the least square algorithm in calibration accuracy and robustness. Lou *et al.* [66] propose a hybrid calibrator by incorporating a differential evolution (DE) algorithm into a Levenberg-Marquardt (LM) for a FANUC M710ic/50 industrial robot calibration. Experimental results show that the proposed calibrator can reduce the positioning error of the robot from 0.994mm to 0.262mm. Chen *et al.* [67] present an improved beetle swarm optimization (BSO) algorithm to identify an industrial robot's kinematic parameters, achieving a huge reduction of positioning error from 2.95mm to 0.2mm. Mao *et al.* [68] approximate the minimax issue of the robot calibration as a sequential quadratic programming (SQP) issue. Then, the SQP issue is addressed with a primal-dual sub-gradient algorithm to achieve a fast convergence rate. Numerous experimental studies conducted on an ABB IRB 14000 robot demonstrate the proposed method can successfully enhance its calibration accuracy.

In general, the above-mentioned algorithms enable the enhancement of the robot positioning accuracy to a large extent. However, they frequently encounter issues like local optimum, poor stability, slow convergence rate, etc., which damage the calibration performance of the robots. Hence, this paper presents a novel algorithm that combines an adaptive and momental bound algorithm with decoupled weight decay (AdaModW). It contains the following ideas:

a) Adopting an adaptive moment estimation (Adam) algorithm to achieve a high convergence rate.
b) Introducing a hyperparameter into the Adam algorithm to define the length of memory, effectively addressing the issue of the abnormal learning rate.
c) Interpolating a weight decay coefficient to address the inadequacy of weight decay and improve its generalization.

Key contributions of this paper are below:
a) An efficient AdaModW algorithm. It interpolates a learning rate control hyperparameter and a weight decay coefficient into an Adam algorithm, achieving a high convergence rate and calibration accuracy.
b) Empirical studies conducted on an HRS-JR680 industrial robot demonstrate the superiority of the AdaModW algorithm in calibration accuracy compared with state-of-the-art algorithms.

Section II briefly describes the preliminaries. The proposed methodology is detailed in Section III. Section IV discusses the experimental results. Section V concludes the paper and gives several future works.

## II. PRELIMINARIES

The most classical kinematics model, i.e., the D-H model, consists of four parameters, i.e., link twist angle ($\alpha_i$), joint angle ($\theta_i$), link length ($a_i$), and $d_i$ (link offset) [1]. Table I lists the nominal D-H parameters of the adopted robot. The transformation matrix of the D-H model is given as:

$$\Gamma_i = \begin{bmatrix} \cos\theta_i & -\sin\theta_i \cos\alpha_i & \sin\theta_i \sin\alpha_i & a_i \cos\theta_i \\ \sin\theta_i & \cos\theta_i \cos\alpha_i & -\cos\theta_i \sin\alpha_i & a_i \sin\theta_i \\ 0 & \sin\alpha_i & \cos\alpha_i & d_i \\ 0 & 0 & 0 & 1 \end{bmatrix}, \quad (1)$$

Then, the end-effector of the robot is described as:

$$\Gamma = \Gamma_1\Gamma_2\Gamma_3\cdots\Gamma_6. \tag{2}$$

The kinematic model deviation can be represented as:

$$d\Gamma = \begin{bmatrix} \Delta H & \Delta O \\ 0 & 1 \end{bmatrix}, \tag{3}$$

where $\Delta H$ is the rotation matrix error of the robot, and $\Delta O$ is the position error of the robot. Note that the holomorphic differential of $\Gamma_i$ can be described as:

$$d\Gamma_i = \frac{\partial \Gamma_i}{\partial \alpha_i}\Delta\alpha_i + \frac{\partial \Gamma_i}{\partial \theta_i}\Delta\theta_i + \frac{\partial \Gamma_i}{\partial a_i}\Delta a_i + \frac{\partial \Gamma_i}{\partial d_i}\Delta d_i. \tag{4}$$

According to (2), (3) and (4), the pose error model is given as:

$$Error = \begin{bmatrix} U_1 & U_2 & U_3 & U_4 \end{bmatrix} \begin{bmatrix} \Delta\alpha \\ \Delta a \\ \Delta d \\ \Delta\theta \end{bmatrix} = U\Delta g, \tag{5}$$

where $U$ refers to the Jacobian matrix, and $\Delta g$ is the D-H parameter deviations vector. Note that the dimension of the D-H parameter vector is set as 24, i.e., $o=24$.

## III. KINEMATIC PARAMETERS IDENTIFICATION

### A. AdaModW Algorithm

In this part, we introduce the AdaModW algorithm and its unique features. Initially, we set the dynamic upper bound to avoid the adaptive learning rate increasing too fast and surpassing the historically largest one [69]. This method significantly enhances the stability of the model and ensures equilibrium in the learning process. Moreover, this paper decreases model complexity and avoids overfitting by decoupling weight decay, thereby enhancing its generalizability [70]. In contrast to $L_2$ regularization, weight decay operates directly on the parameter updates, which penalizes large weights and maintains optimized performance. The specific details of the AdaModW algorithm are as follows:

**a) First-order moment estimate.** Calculate the exponential moving average of the gradient:

$$m_t = \beta_1 m_{t-1} + (1-\beta_1)U_t, \tag{6}$$

where $m_t$ represents the first-order moment estimate at the $t$-th iteration, $U_t$ is the gradient matrix at the $t$-th iteration (the Jacobian matrix in this paper), and $\beta_1$ is the decay weight of the first-order moment estimate.

**b) Second-order moment estimate.** Calculate the exponential moving average of the squared gradient:

$$z_t = \beta_2 z_{t-1} + (1-\beta_2)U_t^2, \tag{7}$$

where $z_t$ denotes the second-order moment estimate at the $t$-th iteration, $U_t^2$ is the square of the gradient at the $t$-th iteration, $\beta_2$ is the decay weight of the second-order moment estimate.

**c) Biases calibration.** According to the literature [30], $m$ and $v$ are inclined to zero at the beginning of training, causing the distance estimate to be biased towards zero. Consequently, the corrections for the biases are required, which are formulated as:

$$\hat{m}_t = \frac{m_t}{1-\beta_{1,t}}, \hat{z}_t = \frac{z_t}{1-\beta_{2,t}}, \tag{8}$$

where $\hat{m}_t$ and $\hat{z}_t$ denote the corrected $m_t$ and $z_t$, respectively.

**d) Update rules.** The D-H parameters are updated as:

$$g_t = g_{t-1} - \frac{\eta}{\sqrt{\hat{z}_t}+\sigma}\hat{m}_t, \tag{9}$$

where $\eta$ is the learning rate, and $\sigma$ is a tiny constant that prevents division by zero. Then, this work defines the adaptive learning rate as:

$$\kappa_t = \frac{\eta_t}{\sqrt{\hat{z}_t}+\sigma}, \tag{10}$$

**e) Adjusting adaptive learning rate.** Motivated by the exponential moving average, we update the adaptive learning rate by computing its average. Specifically, it consists of the following operations:

$$b_t = \beta_3 b_{t-1} + (1-\beta_3)\kappa_t, \tag{11}$$

where $b_t$ is the adjusted value at the $t$-th iteration, and $\beta_3$ denotes a discount factor for controlling $b_t$. Notably, the current $b_t$ represents an interpolation between the current $\kappa_t$ and the prior $b_{t-1}$. The data's mean range in the exponential moving average is $1/\beta_3$, as deduced from its expanded form. For instance, with $\beta_3$ set to 0.9, 0.99, and 0.999, the average range spans 10, 100, and 1000 periods, respectively. The other settings for $\beta_3$ follow the same logic.

Note that, (16) has another expression, which can describe $b_t$ as an exponential weighted moving average. Thus, $b_t$ can also be formulated as:

$$b_t = (1-\beta_3)\left[b_{t-1} + \beta_3 b_{t-2} + \beta_3^2 b_{t-3} + \cdots + \beta_3^{t-1} b_0\right]. \tag{12}$$

Based on (17), $b_t$ retains a 'long-term memory' of the set $\{b_{t-1}, ..., 0\}$. Notably, $b_0$ is set to 0 and is not subject to bias correction.

Then, to prevent excessively high learning rates, we update the adaptive learning rate through a bounding operation, i.e., by comparing it with the current adjusted value $b_t$. Hence, it can be reformulated as:

$$\hat{\kappa}_t = \min(\kappa_t, b_t), \tag{13}$$

where $\hat{\kappa}_t$ represents the ultimate learning rate calculated by bounding operation. Hence, the training process can effectively constrain the output by the adjusted value. Then, (14) can be rewritten as:

$$g_t = g_{t-1} - \hat{\kappa}_t \hat{m}_t, \tag{14}$$

e) **Decoupled weight decay strategy.** It is to implement weight decay directly on the parameter updates. Hence, (19) is updated as:

$$g_t = g_{t-1} - \hat{\kappa}_t (\hat{m}_t + \xi g_{t-1}), \tag{15}$$

where $\xi$ denotes the weight decay coefficient.

IV. EXPERIMENTS AND RESULTS

*A. General Settings*

**Evaluation Metrics.** To validate the experiments, three evaluation metrics, i.e., mean square error (RMSE), mean error (MEAN), and maximum error (MAX) are employed, which can be given as [1], [71]-[78]:

$$MAX = max(|C_i - \hat{C}_i(g)|), i = 1, 2, ..., n,$$

$$MEAN = \frac{1}{n}\sum_{i=1}^{n}|C_i - \hat{C}_i(g)|,$$

$$RMSE = \sum_{i=1}^{n}\left(\sqrt{\frac{1}{n}(C_i - \hat{C}_i(g))^2}\right).$$

where $n$ refers to the sample count, $C_i$ is the measured wire length, and $\hat{C}_i(h)$ represents the nominal wire length

**Experimental Setup.** It contains an HSR-JR680 industrial robot, a dedicated robot controller, a cable-extension transducer, a cable-extension indicator, and a computer.

**Experimental Process.** Firstly, 2000 sample points are selected in the robot's workspace and measured with a cable-extension transducer. Note that the collected sample points are evenly distributed across the entire workspace of the robot. Afterwards, the proposed AdaModW algorithm is adopted to optimize the D-H parameter vector by training the selected samples, aiming to greatly decrease the robot positioning error.

*B. Comparisons*

**Compared results.** In this part, this paper conducts several experiments to verify the performance of the proposed algorithm. Table II details the compared algorithms. Table III illustrates the calibration accuracy before calibration and after calibration by M1-8. Table IV shows the iteration count and computational efficiency of M1-8. Fig. 2 depicts the experimental results in the calibration accuracy of M1-8. Fig. 3 shows the training processes and total time costs of M1-8. Fig. 4 illustrates the positioning accuracy of all algorithms. The deviations of the kinematic parameters are recorded in Table V. From the above results, we obtain the following findings:

a) **In the realm of calibration accuracy, AdaModW surpasses other algorithms.** As shown in Table III and Fig. 2, it is evidently seen that M8, i.e., the presented algorithm, significantly advances robot calibration accuracy, demonstrating a superior performance advantage over other cutting-edge algorithms. For instance, the RMSE achieved by M8 is 0.646mm, which is 23% less than M1's 0.839mm, 53.36% less than M2's 1.385mm, 21.79% less than M3's 0.826mm, 5.28% less than M4's 0.682mm, 1.07% less than M5's 1.07mm, 4.86% less than M6's 0.679mm, 14.55% less than M7's 0.756mm. Considering MEAN, the outputs by M1-8 are 0.739mm, 1.288mm, 0.713mm, 0.59mm, 0.562mm, 0.588mm, 0.662mm, and 0.522mm, respectively. Hence, M1's MEAN is also considerably lesser than that of its counterparts. Moreover, M8's MAX is 1.011mm, which is 38.17%, 61.46%, 36.25%, 36.25%, 13.07%, 2.03%, 10.13%, and 24.33% than that of M1-7.

b) **AdaModW boasts a remarkably rapid convergence rate.** For instance, from Table IV and Fig. 3, M1-7 require 58, 13, 82, 60, 68, 22, and 15 iterations, respectively, to achieve convergence in RMSE. However, M8 converges to RMSE within only 10 iterations, far fewer than its peers. The high convergence rate may be attributed to the innovative learning scheme introduced in Section III.

c) **AdaModW's computational efficiency is competitive.** For instance, From Table IV, M8 takes 10.3 seconds to converge in RMSE, which is 70.82%, 11.21%, 50.72%, 68.88%, 67.81%, 77.75%, and 36.02% lower than that of M1-9, respectively. Thus, M8 achieves a competitive level of computational efficiency.

TABLE I. HRS JR680 INDUSTRIAL ROBOT D-H PARAMETERS.

| Joint $i$ | 1 | 2 | 3 | 4 | 5 | 6 |
|---|---|---|---|---|---|---|
| $α_i/°$ | -90 | 0 | -90 | 90 | -90 | 0 |
| $a_i/mm$ | 250 | 900 | -205 | 0 | 0 | 0 |
| $d_i/mm$ | 653.5 | 0 | 0 | 1030.2 | 0 | 200.6 |
| $θ_i/°$ | 0 | -90 | 180 | 0 | 90 | 0 |

TABLE II. COMPARED ALGORITHMS.

| No. | Description |
|---|---|
| M1 | The Genetic Algorithm (GA) is introduced in [71]. |
| M2 | The EKF algorithm in [72]. |
| M3 | The BSO algorithm is introduced in [67]. |
| M4 | The LM algorithm [73] is frequently employed for robot calibration, which can effectively address the issue of over-fitting. |
| M5 | The improved whale optimization algorithm (IWOA) [75] integrates deep search strategies with the whale optimization algorithm (WOA), enhancing its global search capabilities significantly. |
| M6 | Radial basis function neural network (RBFNN) is introduced in [76]. |
| M7 | The Adam algorithm combines the concepts of momentum and adaptive learning rates, ensuring efficient computational performance [79]. |
| M8 | The proposed AadModW algorithm. |

TABLE III. THE CALIBRATION ACCURACY BEFORE CALIBRATION AND AFTER CALIBRATION BY M1-8.

| Models | RMSE/mm | MEAN/mm | MAX/mm |
|---|---|---|---|
| BC | 4.56 | 4.45 | 5.81 |
| M1 | 0.839±7.3E-3 | 0.739±8.1E-3 | 1.635±8.2E-2 |
| M2 | 1.385±1.1E-2 | 1.288±1.2E-2 | 2.623±1.0E-2 |
| M3 | 0.826±1.3E-2 | 0.713±2.2E-2 | 1.586±2.1E-2 |
| M4 | 0.682±2.1E-2 | 0.590±2.1E-2 | 1.163±2.1E-2 |
| M5 | 0.653±2.0E-2 | 0.562±1.8E-2 | 1.032±1.2E-2 |
| M6 | 0.679±1.0E-2 | 0.588±1.2E-2 | 1.125±9.6E-3 |
| M7 | 0.756±1.1E-2 | 0.662±8.2E-3 | 1.336±7.9E-3 |
| M8 | 0.646±1.2E-2 | 0.552±1.1E-2 | 1.011±9.2E-3 |

TABLE IV. THE TIME COSTS AND ITERATION COUNTS OF M1-8.

| Models | Iterations (RMSE) | Time/s (RMSE) |
|---|---|---|
| M1 | 58 | 35.3±2.11 |
| M2 | 13 | 11.6±1.63 |
| M3 | 82 | 20.9±2.76 |
| M4 | 60 | 33.1±0.91 |
| M5 | 68 | 32.0±1.07 |
| M6 | 22 | 46.3±2.36 |
| M7 | 15 | 16.1±1.74 |
| M8 | 10 | 10.3±1.5 |

TABLE V. D-H PARAMETERS AFTER M9 CALIBRATION.

| Joint $i$ | $α_i/°$ | $a_i/mm$ | $d_i/mm$ | $θ_i/°$ |
|---|---|---|---|---|
| 1 | -90.3353 | 249.5765 | 654.2351 | 0.5676 |
| 2 | -0.2354 | 899.1245 | -0.2352 | -90.2354 |
| 3 | 90.4562 | -205.0254 | -0.6854 | 180.4565 |
| 4 | -90.2541 | 0.4568 | 1032.1235 | 0.1242 |
| 5 | 90.7654 | -0.1235 | 0.1235 | 90.1215 |
| 6 | 0.1245 | -0.1235 | 199.5464 | 0.7668 |

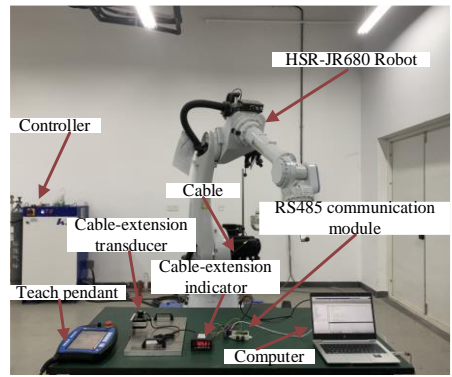

Fig. 1. Experimental system.

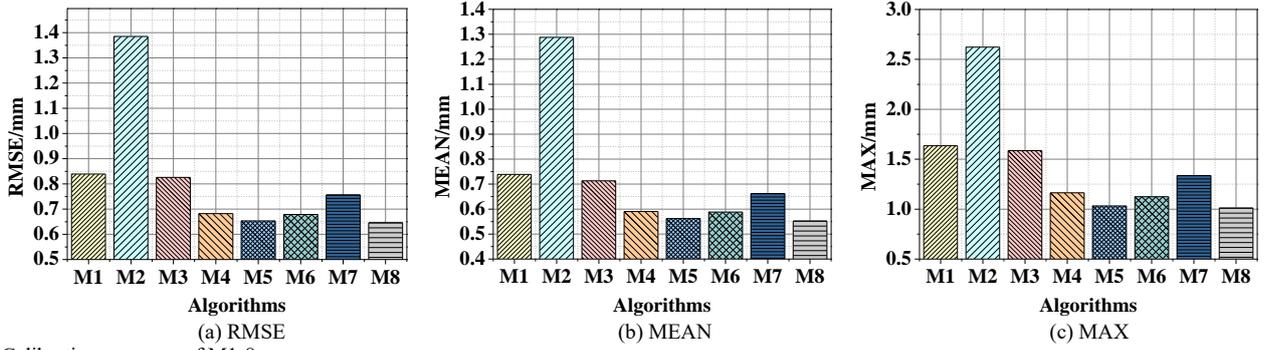

(a) RMSE  (b) MEAN  (c) MAX

Fig. 2. Calibration accuracy of M1-8.

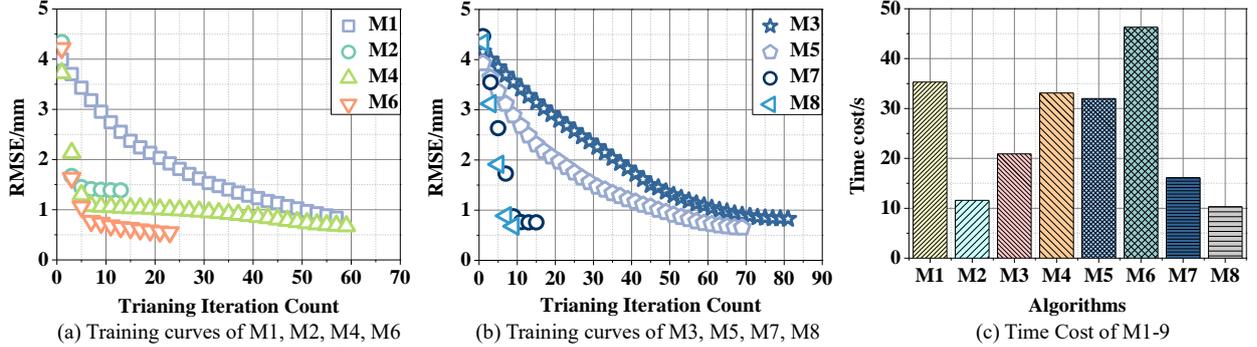

(a) Training curves of M1, M2, M4, M6  (b) Training curves of M3, M5, M7, M8  (c) Time Cost of M1-9

Fig. 3. Training curves and total time cost of M1-8.

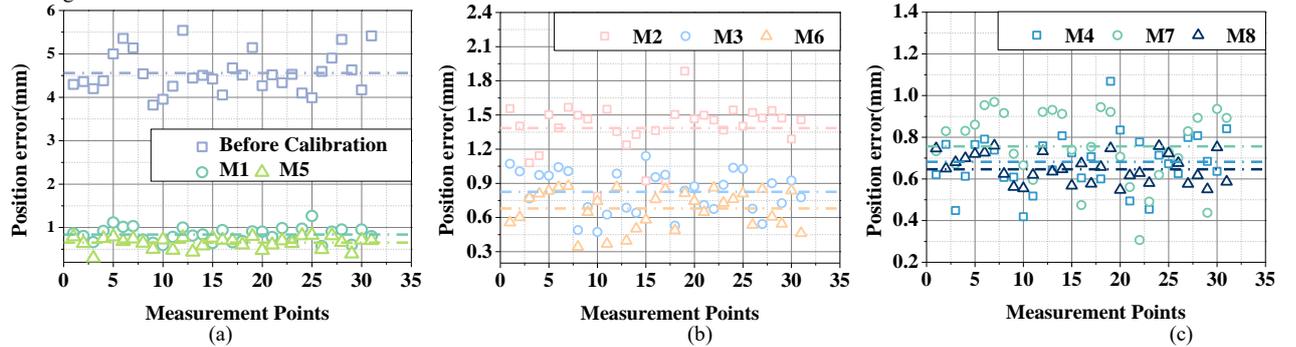

(a)  (b)  (c)

Fig. 4. The positioning accuracy of the robot after calibration through all compared algorithms. Notably, the dotted lines indicate the mean values. It shows that M8 achieves the highest positioning accuracy.

d) **AdaModW can enhance positioning accuracy greatly.** In this paper, 30 data points are randomly chosen from the tested datasets for a comparative analysis of position errors across M1 to M8. As shown in Fig. 4, all models are capable of substantially lowering the positioning errors in robots, among which M8 performs the best. Additionally, the deviations of the kinematic parameters optimized by M8 are acceptable and kept within limits, indicating that the accuracy gains are not attributable to overfitting.

## I. Conclusions

Aiming to achieve efficient robot calibration, this paper proposes a novel AdaModW algorithm, which introduces a hyperparameter into the Adam algorithm to define the length of memory, effectively addressing the issue of the abnormal learning rate. Moreover, it interpolates a weight decay coefficient to address the inadequacy of weight decay and improve its generalization. Empirical studies on the robots show that the proposed algorithm has high calibration accuracy and convergence rate, presenting a practical approach for the field of robot calibration research. Presently, two appealing issues remain open:
a) The robotic accuracy can be further increased by reducing non-geometric errors [80].
b) GPU can be adopted to further enhance the computational efficiency of the AdaModW algorithm [81].